\documentclass[10pt,conference]{IEEEtran}
\usepackage{cite}
\usepackage{amsmath,amssymb,amsfonts}
\usepackage{algorithmic}
\usepackage{graphicx}
\usepackage{textcomp}
\usepackage{listings}
\usepackage{xcolor}
\usepackage{booktabs}
\usepackage{xspace}
\usepackage[most]{tcolorbox}
\usepackage{booktabs}  % For professional table formatting
\usepackage{graphicx}  % For including graphics, if needed
\usepackage{adjustbox} % For table scaling, if needed
\usepackage{multirow} 
\usepackage{hyperref}
\usepackage{url}
\usepackage[all]{nowidow}

\def\BibTeX{{\rm B\kern-.05em{\sc i\kern-.025em b}\kern-.08em
    T\kern-.1667em\lower.7ex\hbox{E}\kern-.125emX}}
\begin{document}
\newcommand{\tool}{CoCoNUT\xspace}
\newcommand{\etal}{et al.\xspace}

\newcommand{\ccot}{CoT}
\newcommand{\basic}{Direct\xspace}
\newcommand{\ovsim}{Similarity\xspace}
\newcommand{\ovsimshort}{Sim\xspace}

\newcommand{\humaneval}{HumanEval\xspace}

\newcommand{\humanevaltasks}{161\xspace}
\newcommand{\advancedtasks}{124\xspace}
\newcommand{\totaltasks}{285\xspace}

\newcommand{\totalllms}{7\xspace}

\newcommand{\hetrace}{HumanEval-Trace\xspace}
\newcommand{\advtrace}{Advanced-Trace\xspace}

\title{\tool: Structural Code Understanding does not fall out of a tree}

\author{\IEEEauthorblockN{Claas Beger}
\IEEEauthorblockA{\textit{Department of Computer Science} \\
\textit{Cornell University}\\
Ithaca, New York \\
cbb89@cornell.edu}
\and
\IEEEauthorblockN{Saikat Dutta}
\IEEEauthorblockA{\textit{Department of Computer Science} \\
\textit{Cornell University}\\
Ithaca, New York}
saikatd@cornell.edu}

\maketitle
\newcommand{\todo}[1][To be updated later!]{{\color{red} TODO: {#1}}}
\definecolor{bgcolor}{rgb}{0.95,0.95,0.95} % Background color for the code
\newcommand{\removed}[1]{\textcolor{red}{#1}}   % Red color for removals
\newcommand{\added}[1]{\textcolor{green}{#1}}   % Green color for additions

\definecolor{WowColor}{rgb}{.75,0,.75}
\definecolor{SubtleColor}{rgb}{0,0,.50}
\newcommand{\cmark}{\ding{51}\xspace}%
\newcommand{\xmark}{\ding{55}\xspace}%
\newcommand{\lh}[1]{\textcolor{blue}{(Linghao: #1)}}
\newcommand{\vp}[1]{\textcolor{blue}{(Vishesh: #1)}}

% general
%\renewcommand{\comment}[1]{}
\ifdefined\Comment
        \renewcommand{\Comment}[1]{}
\else
        \newcommand{\Comment}[1]{}
\fi
% inline
\newcommand{\NA}[1]{\textcolor{SubtleColor}{ {\tiny \bf ($\star$)} #1}}
\newcommand{\Fix}[1]{\textcolor{red}{[#1]}}
\newcommand{\MN}[1]{\textcolor{blue}{#1}}
\newcommand{\LATER}[1]{\textcolor{SubtleColor}{#1}}
\newcommand{\TBD}[1]{\textcolor{SubtleColor}{ {\tiny \bf (!)} #1}}
\newcommand{\PROBLEM}[1]{\textcolor{WowColor}{ {\bf (!!)} {\bf #1}}}

% as margin notes
\newcounter{margincounter}
\newcommand{\displaycounter}{{\arabic{margincounter}}}
\newcommand{\incdisplaycounter}{{\stepcounter{margincounter}\arabic{margincounter}}}

\newcommand{\fTBD}[1]{\textcolor{SubtleColor}{$\,^{(\incdisplaycounter)}$}\marginpar{\tiny\textcolor{SubtleColor}{ {\tiny $(\displaycounter)$} #1}}}

\newcommand{\fPROBLEM}[1]{\textcolor{WowColor}{$\,^{((\incdisplaycounter))}$}\marginpar{\tiny\textcolor{WowColor}{ {\bf $\mathbf{((\displaycounter))}$} {\bf #1}}}}

\newcommand{\fLATER}[1]{\textcolor{SubtleColor}{$\,^{(\incdisplaycounter\dagger)}$}\marginpar{\tiny\textcolor{SubtleColor}{ {\tiny $(\displaycounter\dagger)$} #1}}}

\newcommand{\mypara}[1]{\vspace{.03in}\noindent \textbf{#1.}}
\

\lstset{
    language=Python,
    basicstyle=\ttfamily\footnotesize, % Font style and size
    numbers=left, % Line numbers on the left
    numberstyle=\tiny\color{gray}, % Style of line numbers
    stepnumber=1, % Line number interval
    numbersep=8pt, % Distance between line numbers and code
    backgroundcolor=\color{white}, % Background color
    showspaces=false,
    showstringspaces=false,
    showtabs=false,
    frame=single, % Frame around the code
    rulecolor=\color{black},
    tabsize=4, % Tab size
    captionpos=b, % Position of the caption (b: bottom)
    breaklines=true, % Automatic line breaking
    breakatwhitespace=false,
    keywordstyle=\color{blue}, % Keyword color
    commentstyle=\color{green!60!black}, % Comment color
    stringstyle=\color{orange}, % String color
    morekeywords={as} % Additional keywords
}

\newcommand{\lstbg}[3][0pt]{{\fboxsep#1\colorbox{#2}{\strut #3}}}
\lstdefinelanguage{diff}{
    language=Python,
    basicstyle=\small\ttfamily,
    morecomment=[f][\lstbg{red!20}]-,
    morecomment=[f][\lstbg{green!20}]+,
    morecomment=[f][\textit]{@@},
    numbers=left, % Line numbers on the left
    numberstyle=\tiny\color{gray}, % Style of line numbers
    stepnumber=1, % Line number interval
    numbersep=8pt, % Distance between line numbers and code
    backgroundcolor=\color{white}, % Background color
    showspaces=false,
    showstringspaces=false,
    showtabs=false,
    frame=single, % Frame around the code
    rulecolor=\color{black},
    tabsize=4, % Tab size
    captionpos=b, % Position of the caption (b: bottom)
    breaklines=true, % Automatic line breaking
    breakatwhitespace=false,
    keywordstyle=\color{blue}, % Keyword color
    commentstyle=\color{green!60!black}, % Comment color
    stringstyle=\color{orange}, % String color
    morekeywords={as,self} % Additional keywords
    keepspaces=true,
    columns=fullflexible,
  %morecomment=[f][\textit]{---},
  %morecomment=[f][\textit]{+++},
}

\definecolor{ghgreen}{rgb}{0.90,1,0.93}
\definecolor{ghred}{rgb}{1,0.88,0.94}

\definecolor{codegreen}{rgb}{0,0.6,0}
\definecolor{codegray}{rgb}{0.5,0.5,0.5}
\definecolor{codepurple}{rgb}{0.58,0,0.82}
\definecolor{backcolour}{rgb}{0.95,0.95,0.92}

\lstset{  
    language=Java,  
    morecomment=[f][\lstbg{red!20}]-,
    morecomment=[f][\lstbg{green!20}]+,
    morecomment=[f][\textit]{@@},
    %backgroundcolor=\color{backcolour},
    commentstyle=\color{codegreen},
    keywordstyle=\color{codepurple},
    numberstyle=\tiny\color{codegray},
    stringstyle=\color{blue},
    basicstyle=\small\ttfamily,
    breakatwhitespace=false,
    breaklines=true,
    captionpos=b,
    keepspaces=true,
    numbers=left,
    numbersep=5pt,
    tabsize=4,
    columns=fullflexible
}

\definecolor{codegreen}{rgb}{0,0.6,0}
\definecolor{codegray}{rgb}{0.5,0.5,0.5}
\definecolor{codepurple}{rgb}{0.58,0,0.82}
\definecolor{backcolour}{rgb}{0.95,0.95,0.92}

\newcommand{\circledsup}[1]{%
  \tikz[baseline=(char.base)]{%
    \node[shape=circle,draw,inner sep=0.5pt] (char) {#1};}%
}

\definecolor{diffadd}{rgb}{0.9,1,0.9}  % Light green
\definecolor{diffdel}{rgb}{1,0.9,0.9}  % Light red

\newcommand{\lessonblock}[1]{
%\stepcounter{mylessoncounter}
\begin{tcolorbox}[
%title=Lessons Learned \themylessoncounter,
colback=black!2!white, colframe=black, boxsep=0.1pt,boxrule=0.1pt]
%\textbf{\themylessoncounter:} 
\emph{#1}
\end{tcolorbox}
}

%\newcounter{mylessoncounter}

\begin{abstract}
Large Language Models (LLMs) have shown impressive performance across a wide array of tasks involving both structured and unstructured textual data. More recently, LLMs have demonstrated remarkable abilities to generate code across different programming languages. Recent results on various benchmarks for code generation, repair, or completion suggest that certain models have programming abilities comparable to or even surpass humans. In this work, we demonstrate that the high performance on such benchmarks does not correlate to humans' innate ability to understand the structural control flow of code.

For this purpose, we extract code solutions from the HumanEval benchmark, which the relevant models perform very strongly on, and trace their execution path using function calls sampled from the respective test set. Using this dataset, we investigate the ability of \totalllms state-of-the-art LLMs to match the execution trace and find that, despite the model's abilities to generate semantically identical code, they possess only limited ability to trace the execution path, especially for longer traces and specific control structures. We find that even the top-performing model, Gemini 1.5 Pro can only fully correctly generate the trace of 47\% of HumanEval tasks. 

In addition, we introduce a subset for three key structures not, or only contained to a limited extent in HumanEval: Recursion, Parallel Processing, and Object Oriented Programming principles, including concepts like Inheritance and Polymorphism. Besides OOP, we show that none of the investigated models achieve an average accuracy of over 5\% on the relevant traces. Aggregating these specialized parts with the ubiquitous HumanEval tasks, we present the Benchmark CoCoNUT: \textbf{Co}de \textbf{Co}ntrol Flow for \textbf{N}avigation \textbf{U}nderstanding and \textbf{T}esting, which measures a models ability to trace the execution of code upon relevant calls, including advanced structural components. 
We conclude that the current generation LLMs still need to significantly improve to enhance their code reasoning abilities. We hope our dataset can help researchers bridge this gap in the near future.
% We show a significant gap in performance in this area, which we hope to address by distributing our dataset.
\end{abstract}

\begin{IEEEkeywords}
Code Understanding, Large Language Models, Code Execution, Benchmarks
\end{IEEEkeywords}

\section{Introduction}
Large Language Models (LLMs), such as GPT-4 and Llama, have demonstrated remarkable progress across diverse tasks and are now widely used in applications like code generation (e.g., GitHub Copilot~\cite{copilot}), healthcare~\cite{clusmann2023future}, and finance~\cite{bloomgpt}.

In programming, LLMs show great promise by automating tasks such as generating code, authoring tests, detecting bugs, and repairing programs. However, achieving true success in these tasks requires LLMs to not only generate code but also \emph{reason} about its behavior. To develop such capabilities, robust benchmarks that evaluate code reasoning are essential.

Current benchmarks like HumanEval~\cite{chen2021evaluating} and MBPP~\cite{austin2021programsynthesislargelanguage} primarily assess code generation for simple tasks and are already saturating, with GPT4 achieving over 90\% accuracy \cite{evalplusEvalPlusLeaderboard}. While newer benchmarks include execution reasoning tasks (e.g. CruxEval\cite{gu2024cruxeval}, LiveCodeBench\cite{jain2024livecodebench}), they do not test the understanding of control flows. Thus, a more structurally focused code reasoning benchmark is needed.

\begin{figure*}[htbp]
\centering
\includegraphics[width=0.7\linewidth]{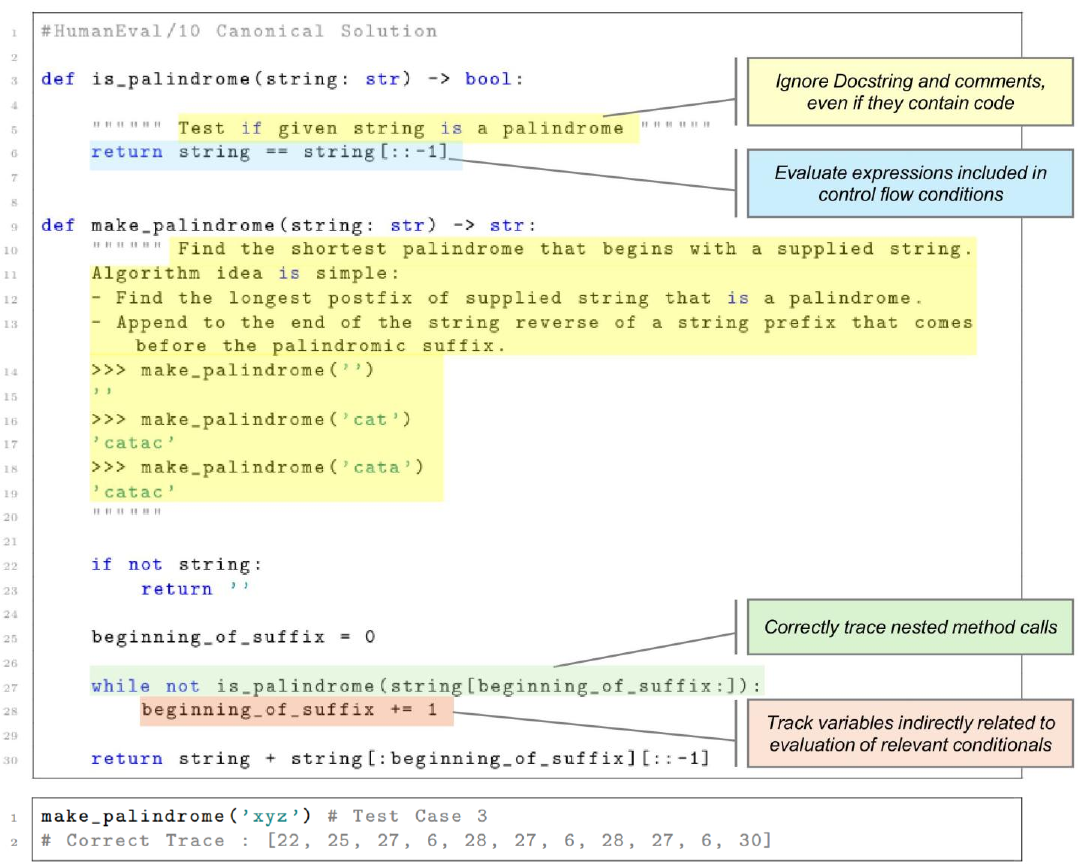}
\caption{Tracing the execution flow requires a diverse understanding of code structures, even for shorter sequences like HumanEval Tasks. We list a couple of examples for Task ID 10.}
\label{fig:NonTrivial}
\end{figure*}

\mypara{Our Work}
To address this, we introduce \tool -- a novel benchmark for evaluating LLMs' ability to trace complex control flows. Each task in \tool includes a program and test input, requiring the model to generate a trace of line numbers executed by the program. This challenging task necessitates handling long-term dependencies, control structures, and nested expressions, as shown in Figure~\ref{fig:NonTrivial}. We argue that tracing execution is a natural test of a model's understanding of code.

Currently, \tool includes \humanevaltasks tasks and 1083 traces derived from HumanEval (\emph{HumanEval-Trace}), and \advancedtasks tasks with 620 traces focusing on advanced programming structures such as recursion, parallel processing, and object-oriented principles (\emph{Advanced-Trace}) -- totaling \totaltasks tasks and 1703~traces.

\looseness=-1
\mypara{Results} We evaluate \totalllms state-of-the-art LLMs, including proprietary models (GPT-4o, Gemini 1.5 Pro, and Claude 3.5 Sonnet) and open-source models (e.g., LLama3.1, Qwen2.5Coder, Mistral Codestral 22B, and CodeLLama 7B). While LLMs excel at tracing simple programs, their performance on advanced control flows lags far behind their code generation abilities. For instance, the top-performing model, Gemini 1.5 Pro, traces only 47\% of HumanEval tasks, and accuracy sharply declines for traces longer than 25 lines. Surprisingly, advanced prompts like Chain-of-Thought offer limited benefits and sometimes degrade the performance on complex tasks. We provide a detailed discussion of results in \S~\ref{sec:results}.
Our code and dataset are available 
 \href{https://github.com/ClaasBeger/StructuralCodeUnderstanding}{here}.

\section{\tool Framework}
\subsection{Tasks} \label{Tasks}
We utilize the HumanEval dataset proposed by \cite{chen2021evaluatinglargelanguagemodels}, which consists of 164 Python programming tasks and serves as a suitable benchmark for structural understanding. These tasks are frequently used to evaluate LLMs, ensuring a correlation between control flow understanding and \linebreak task-solving performance. 

To adapt the dataset, we merge the function signature (including docstring examples) with the canonical solution. Using an abstract syntax tree, we parse the testing code to identify all calls to the candidate function and extract their arguments. HumanEval tasks are concise, with function bodies ranging from 1 to 28 lines and an average of 7.7 tests per function, resulting in approximately 1250 potential execution traces. We employ the Python tracer provided by the sys library to analyze execution for each argument set provided by a test call, filtering out sets requiring more than 1024 consecutive tokens, yielding 1083 argument sets across 161~tasks.

Additionally, Liu et al. introduced EvalPlus \cite{liu2023is}, which expands the number of tests for HumanEval tasks, revealing that model performance was previously overestimated. While we generate an additional dataset based on EvalPlus (over 120,000 traces without trace length filtering), we limit our scope to a preliminary evaluation on the base HumanEval tests.

\subsection{Evaluation Metrics}
We employ two main evaluation metrics: \emph{Exact Match} and \emph{Similarity}. Exact Match provides a straightforward measure of whether a model can fully trace the relevant execution. To avoid bias from tasks with varying numbers of tests, we compute task-level accuracy first, then aggregate it as an overall \emph{Accuracy Mean}. Note that this metric does not apply when evaluating traces of different lengths, as trace lengths can vary significantly and are not always tied to code \linebreak line numbers.

While accuracy is intuitive, it does not capture the diversity of errors. A trace that deviates by only one line is better than a completely incorrect prediction. To address this, we use a \emph{Similarity} metric based on the Gestalt-pattern matching algorithm~\cite{ratcliff1988pattern}. This metric focuses on matching contiguous subsequences, emphasizing longer matches over distributed ones. The similarity between two sequences \( A \) and \( B \) is computed as:
\begin{equation}
S(A, B) = \frac{2 \cdot M}{|A| + |B|}
\end{equation}
where \( M \) is the total number of matching characters in contiguous blocks, and \( |A| \) and \( |B| \) are the lengths of the sequences. The metric ranges from 0 to 1 and complements the accuracy by quantifying partial matches. For incorrectly generated traces, we also report \emph{False Similarity}.
Further, we distinguish the hard accuracy (\textbf{Acc Hard}), as the fraction of tasks that the models solve completely, meaning they are able to reproduce the trace of all given tests.

For concurrency, traditional metrics are insufficient due to variations in execution order across threads. We isolate concurrent segments, compare overlapping indices, and sort the data before applying the similarity metric. While lenient, this approach requires models to correctly identify the start and end of concurrent execution as a prerequisite, which we consider a valid trade-off.

\section{Methodology}
\subsection{Research Questions}
In our work, we are particularly interested in the following research questions:
\begin{itemize}
    \item\textbf{RQ1:} How does the understanding of execution semantics relate to code generation abilities? 
    \item\textbf{RQ2:} What is the impact of different prompting techniques on the overall performance?
    \item\textbf{RQ3:} How well can current state-of-the-art language models understand the execution semantics of programs? How does this ability change with the increasing complexity of programs, as measured by length or by introducing more advanced programming concepts?
\end{itemize}

\subsection{Dataset Curation}
As described earlier, we collect programs from multiple sources, including HumanEval. For the advanced topics, we extract LeetCode programs based on tags from multiple open-source repositories (\cite{walkcccLeetCodeSolutions}, \cite{githubGitHubNeetcodeghleetcode}). Further, we complement this data by collecting programs from the multi-language programming platform RosettaCode\cite{rosettacodeRosettaCode}, which also provides relevant topic tags. We extract the actual programs from a related open-source repository~\cite{githubRosettaCodeDataTaskMain}. Unfortunately, these programs often do not have corresponding tests or relevant execution logic. We employ GPT4o to generate simple base test cases which we expand manually. Generally, we create five tests for every sample and try to employ increasing difficulty (regarding number of calls, trace length, complex control flows and similar aspects), which is necessary to invoke the topical functionality. Since the advanced programs generally feature multiple components (methods, classes, etc.), we create a main code block for the models to trace, rather than a single method call. For the topic of concurrency, as briefly mentioned in the previous section, we need to make further adjustments to enable valid tracing. In our prompt, we describe how the model may use parentheses to mark concurrent execution, We further demonstrate it in the given prompt and tolerate smaller deviations like \linebreak multiple parentheses.\newline

We find that models initially struggle to find the correct format, so we employ one-shot prompting, giving the model a very simple code call with the corresponding execution trace to demonstrate the correct format. We explicitly ask the model to refrain from producing other output in order to produce a meaningful comparison with Chain-of-Thought prompting. We observe that especially the smaller LLama model, but also some of the larger models deviate from this instruction sometimes, which we score as a zero in both metrics. In the given prompt, we merge the function signature and docstring with the canonical solution and annotate the lines with an index number to alleviate potential ambiguity through newlines or comments. Separate from the given code, we name the called function and the argument list. If necessary, we provide superficial cleaning of the predictions (removing whitespaces, markdown, or Python annotations). If not specified otherwise, we use greedy decoding and a maximum token length of 1024 for direct prompting, as well as 4096 for CoT. We elect to increase the token count in this manner to avoid cutting off the generation before the final trace is reached. 

\section{Results}
\label{sec:results}
\begin{table*}[!htb]
\centering
\caption{Performance Overview on the HumanEval-Trace}  % Center the caption
\label{tab:task_performance}
\begin{tabular}{@{}lcccccccc@{}}
\toprule
\multirow{2}{*}{\textbf{Model}} & \multicolumn{4}{c}{\textbf{\ccot}} & \multicolumn{4}{c}{\textbf{\basic}} \\ \cmidrule(r){2-5} \cmidrule(l){6-9}
 & \textbf{Acc Hard (\%)} & \textbf{Acc Mean (\%)} & \textbf{\ovsimshort} & \textbf{False \ovsimshort} & \textbf{Acc Hard (\%)} & \textbf{Acc Mean (\%)} & \textbf{\ovsimshort} & \textbf{False \ovsimshort} \\ \midrule
Gemini1.5-Pro 002 & \textbf{47.2} & \textbf{66.2} & \textbf{0.88} & 0.37 & \textbf{47.0} & \textbf{65.7} & \textbf{0.89} & 0.37 \\
Claude3.5-Sonnet & 41.0 & 61.6 & 0.87 & 0.43 & 41.0 & 58.7 & 0.88 & 0.44 \\
GPT4o & 16.8 & 39.4 & 0.75 & 0.50 & 21.2 & 38.8 & 0.75 & 0.50 \\
LLama3.1 70B & 16.2 & 38.1 & 0.76 & 0.52 & 25.5 & 36.0 & 0.71 & 0.42 \\
CodeLLama 34B & 1.2 & 7.6 & 0.46 & 0.43 & 2.5 & 10.0 & 0.57 & 0.52 \\ 
Qwen2.5-Coder 32B & 26.1 & 44.3 & 0.81 & 0.50 & 32.7 & 42.4 & 0.78 & 0.44 \\
Codestral 22B & 9.3 & 25.0 & 0.71 & \textbf{0.57} & 3.1 & 17.8 & 0.66 & \textbf{0.59} \\
LLama3.1 8B & 1.9 & 12.6 & 0.56 & 0.51 & 0.6 & 10.4 & 0.53 & 0.48 \\
Qwen2.5-Coder 7B & 1.9 & 11.0 & 0.61 & 0.56 & 0.0 & 4.1 & 0.56 & 0.55 \\
CodeLLama 7B & 0.0 & 0.1 & 0.28 & 0.28 & 0.0 & 0.0 & 0.41 & 0.41 \\\bottomrule
\end{tabular}
\end{table*}

\begin{figure}[!htb]
    \centering
    \includegraphics[width=\columnwidth]{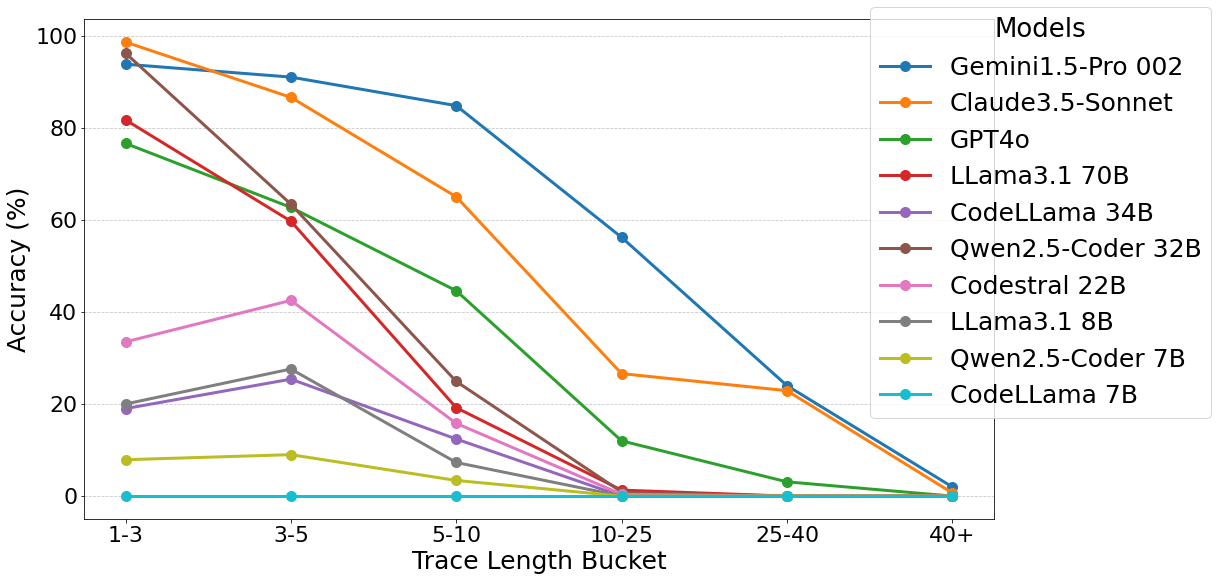}
    \caption{Model Accuracy by Buckets of Trace Length using Direct Prompting.}
    \label{fig:TraceBuckets}
\end{figure}

\subsection{Performance of SOTA LLMs on \tool}
We query three popular large-scale language models that perform at the top of the HumanEval and EvalPlus benchmarks respectively. Namely, we use Claude 3.5 - Sonnet, Gemini Pro 1.5 002, and GPT4o. Since there is relevant work describing that Chain-of-Thought improves execution tracing~\cite{ni2024nextteachinglargelanguage, lyu2024largelanguagemodelscode} (although different from our definition of trace), we specifically look at the performance of direct and Chain-of-Thought prompting. We present the aggregated results of our experiments on HumanEval in Table~\ref{tab:task_performance}. Across all tables, we refer to Similarity as \textbf{Sim} and Accuracy as \textbf{Acc}.

The results show that none of the models are able to match their performance on HumanEval. Further, we observe two interesting points in that the performance comparison between the models does not reflect the ranking on HumanEval, on which GPT4o significantly outranks the other two large models. Upon closer investigation, GPT4o strongly struggles with hallucinations, where function signatures are called directly (which is specified to be incorrect in the prompt), and code lines in the docstrings or unrelated code indices are listed. Further, there is a large difference between the average accuracy over all traces and the number of execution cases that the models are able to fully solve.

\subsection{Correlation on Generative vs Execution Tracing Tasks}
We conduct a small experiment to assess the correlation between the performance in solving the described HumanEval coding problem, as measured by the given test suite, and the generation of execution traces for the provided sample solution. For this experiment, we utilize LLama 3.1 70B, which demonstrated strong performance on HumanEval tasks. We intentionally refrain from using a code-specific model, as its training, potentially influenced by HumanEval, might lead to an overestimation of performance on the code generation task. Applying this model on HumanEval results in a pass rate of 80.4\% using base tests and 74.8\% using evalplus. While we include evalplus tests for completeness, we emphasize that the comparison using base tests is more significant, as these are the tests with which we reproduce the code execution.

To generate a meaningful difficulty ranking, we order the generative results by the number of tests failed by each solution. Similarly, we rank the execution tracing results based on the average solution similarity per task. We compute Spearman’s rank correlation for these orderings, yielding values of -0.05/-0.09 (base and evalplus) for the Chain-of-Thought approach and 0.06/0.01 for the direct approach. Both pairs of figures suggest no significant correlation between the difficulty of code generation and execution tracing. Additionally, we compute the relative overlap of failed tasks between the two approaches. Approximately 20\% of the tasks that failed in execution tracing also failed during code generation for Chain-of-Thought prompting, and 17\% for direct prompting. These findings support our previous assumption that the difficulty in execution tracing primarily arises from the trace length, but also from the inclusion of specific code constructs that are challenging for the model to execute correctly. The latter aspect will be further discussed in \autoref{sec:Qualitative Errors}. However, it does not appear that there is a significant overlap with the difficulty sources of Code Generation.

\noindent
\lessonblock{\textbf{Answer to RQ1}: Thus, we conclude that there is only a very weak intermediate connection between the model performance on Code Generation and Execution Tracing, which does not apply to task-specific difficulty.}

\begin{figure}[!htb]
    \centering
    \includegraphics[width=\columnwidth]{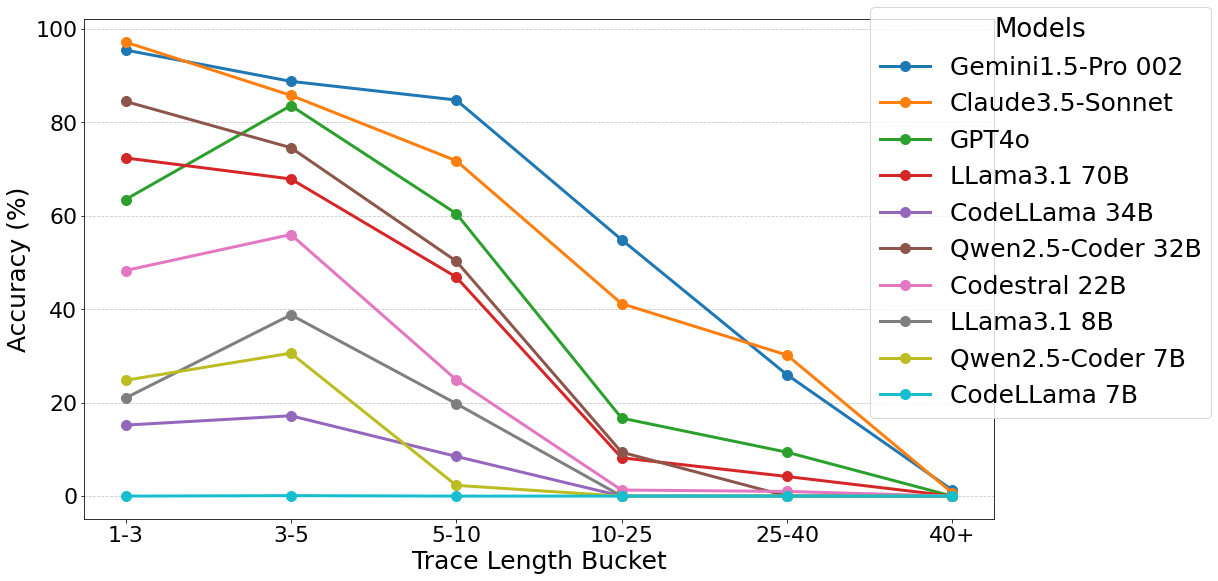}
    \caption{Model Accuracy by Buckets of Trace Length using Chain-of-Thought Prompting.}
    \label{fig:TraceBuckets_cot}
\end{figure}

\subsection{ Direct vs CoT Prompting over Trace Lengths}
Besides the performance metrics, we set out to compare direct and CoT prompting. Our findings suggest that the benefits of CoT may not be as significant for larger models. While Codestral and and the smaller versions of LLama and QwenCoder experience a 3x increase in hard accuracy, Gemini and Claude stay consistent and GPT4o, as well as larger versions of the two previously named models even decreases. Besides the general performance, we were interested in finding how the complexity of the execution, as measured by the length of the code or trace, impacts the performance. Although we find that the length of the execution trace and the code length have a similar effect on the task performance, we generally focus on trace length for evaluation purposes. We distribute the trace lengths in 6 buckets and aggregate performance within them. Our results are shown in \autoref{fig:TraceBuckets} and \autoref{fig:TraceBuckets_cot}.

The stronger models generally start to see a large performance drop from a trace length of 25 onwards. Similarly, medium sized models (70-22B) have a cutoff of around 10 and the smaller models already struggle with lengths beyond 5. This supports the hypothesis that the ability to trace long code executions emerges at scale.  Further, after a trace length of 40, models are generally unable to generate a meaningful trace in the vast majority of cases. For larger traces, some of the models also lost the correct execution path and went into endless loops, which were truncated according to the token limit. We would further like to address the performance of the different code-specific models we included. While some models, like the larger version of QwenCoder seem to have improved capabilities even with smaller parameter counts, others like CodeLLama or CodeStral just perform in line with their parameter count. We hypothesize that this is due to the fact that different models had varying degree of access to relevant execution data during their training period, despite the focus on code. In addition, we observed some errors with instruction-following in some of them.\newline
We also consider the performance using CoT prompting and again observe the most significant impact for smaller models, even though larger models also improve slightly. Notably, Chain-Of-Thought mostly improves the existing capabilities, and models generally tend to lose functionality at the same threshold as direct prompting. This is somewhat surprising since existing works on investigating traces with a focus on variable states tend to claim that CoT naturally combats long-dependence issues through verbose self-documentation~\cite{wei2023chainofthoughtpromptingelicitsreasoning}, similar to alternative approaches like Tree-of-Thought~\cite{yao2023treethoughtsdeliberateproblem} or employment of other means for the model to document intermediate steps \cite{nye2021workscratchpadsintermediatecomputation}. 
%\noindent

\lessonblock{\textbf{Answer to RQ2:} Using our current results, we cannot clearly identify a significantly positive effect of CoT on execution tracing. With regard to our second research question, we conclude that there is some importance to the chosen prompting technique, but the effect varies according to model size and the type of program that is supposed to be traced.}

\begin{table}[!hbt]
\centering
\caption{Performance on a subset of different advanced programming concepts sourced from Open Source projects and competitions. Every program was traced with 5 sample blocks of increasing difficulty.} 
\label{tab:task_performance_adv}
\begin{tabular}{@{}lcc|cc@{}}
\toprule
\multirow{2}{*}{\textbf{Model}} & \multicolumn{2}{c|}{\textbf{CoT}} & \multicolumn{2}{c}{\textbf{Direct}} \\ \cmidrule(r){2-3} \cmidrule(l){4-5}
 & \textbf{Acc Mean (\%)} & \textbf{Sim} & \textbf{Acc Mean (\%)} & \textbf{Sim} \\ \midrule

\multicolumn{5}{l}{\textbf{Object-Oriented} (40 Programs 200 Traces)} \\ \midrule
Gemini1.5-Pro 002 & 14.0 & 0.79 & \textbf{20.0} & \textbf{0.81} \\
Claude3.5-Sonnet & 0.0 & 0.77 & 1.0 & 0.69 \\
GPT4o & 4.5 & \textbf{0.82} & 4.0 & 0.73 \\
LLama3.1 70B& \textbf{15.0} & 0.74 & 10.0 & 0.75 \\
CodeLLama 34B & 0.0 & 0.37 & 0.0 & 0.37 \\
Qwen2.5-Coder 32B& 14.5 & 0.78 & 4.0 & 0.73 \\
Codestral 22B& 1.5 & 0.62 & 1.5 & 0.6 \\
LLama3.1 8B & 0.5 & 0.58 & 1.0 & 0.48 \\
Qwen2.5-Coder 7B & 0.0 & 0.58 & 0.0 & 0.56 \\
CodeLLama 7B & 0.0 & 0.30 & 0.0 & 0.40 \\\midrule
\multicolumn{5}{l}{\textbf{Recursion} (66 Programs 330 Traces)} \\ \midrule
Gemini1.5-Pro 002 & 2.7 & 0.47 & 0.9 & \textbf{0.41} \\
Claude3.5-Sonnet & 0.3 & 0.42 & 1.2 & \textbf{0.41} \\
GPT4o & 2.7 & \textbf{0.49} & \textbf{1.8} & 0.38 \\
LLama3.1 70B& 1.2 & 0.36 & 0.6 & 0.27 \\
CodeLLama 34B & 0.0 & 0.29 & 0.0 & 0.27 \\
Qwen2.5-Coder 32B& \textbf{3.0} & 0.35 & \textbf{1.8} & 0.30 \\
Codestral 22B& 1.0 & 0.29 & 0.0 & 0.29 \\
LLama3.1 8B & 0.3 & 0.21 & 0.0 & 0.35 \\
Qwen2.5-Coder 7B & 0.0 & 0.15 & 0.0 & 0.16 \\
CodeLLama 7B & 0.0 & 0.23 & 0.0 & 0.26 \\
\midrule
\multicolumn{5}{l}{\textbf{Concurrency} (20 Programs 100 Traces)} \\ \midrule
Gemini1.5-Pro 002 & \textbf{1.0} & \textbf{0.41} & \textbf{1.0} & 0.39 \\
Claude3.5-Sonnet & 0.0 & 0.4 & 0.0 & \textbf{0.42} \\
GPT4o & 0.0 & 0.39 & 0.0 & 0.4 \\
LLama3.1 70B& \textbf{1.0} & 0.34 & \textbf{1.0} & 0.33 \\
CodeLLama 34B & 0.0 & 0.24 & 0.0 & 0.27 \\
Qwen2.5-Coder 32B& 0.0 & 0.36 & 0.0 & 0.37 \\
Codestral 22B& 0.0 & 0.29 & 0.0 & 0.38 \\
LLama3.1 8B & 0.0 & 0.28 & 0.0 & 0.26 \\
Qwen2.5-Coder 7B & 0.0 & 0.21 & 0.0 & 0.18 \\
CodeLLama 7B & 0.0 & 0.23 & 0.0 & 0.28 \\ \bottomrule
\end{tabular}
\vspace{0.2cm}
\end{table}

\subsection{Advanced Structural Topics}
\looseness=-1
Besides HumanEval, CoCoNUT employs evaluation on different advanced aspects. We show an overview of the corresponding performance in Table~\ref{tab:task_performance_adv}. Apart from a few models on the task of Object-Oriented Programming Principles, none of the models can provide performance beyond 5\% task accuracy. Further, it is apparent that Chain-Of-Thought only improves performance slightly on Recursion, but not on the other topics. We note that the corresponding programs are generally longer in code, as well as execution traces. However, this is still a very controlled setting since all relevant methods are in a single file and none of the code samples are longer than 200 lines of code. 

We believe that this effectively shows how even SOTA models struggle with fully reproducing the execution calls involving the application of advanced code concepts. More explicitly, we observe that models cannot handle resolving recursion after a certain depth and regularly produce infinite loops. For Concurrency, the models struggle with correctly identifying the start and end of parallel execution, as well as the nested worker calls. For OOP, models generally struggle with navigating the longer code segments needed to set up the necessary class structure and correct call resolution.

\noindent
\lessonblock{\textbf{Answer to RQ3:} We find that state-of-the-art LLMs, despite strong performance on generative tasks, struggle with structural tracing, especially on code featuring diverse sources of complexity. In general, larger models appear to feature a basic understanding of simple control flow structures but traces beyond the length of 40, as well as advanced structural code concepts, remain a significant challenge for all models to reason about.
%make it very difficult for the model to predict the correct trace.
}

\section{Discussion}
\subsection{Qualitative Error Analysis} \label{sec:Qualitative Errors}
\begin{figure*}[!htbp] % Allows LaTeX to place the figure more flexibly
\centering % Center the figure
\includegraphics[width=0.85\linewidth]{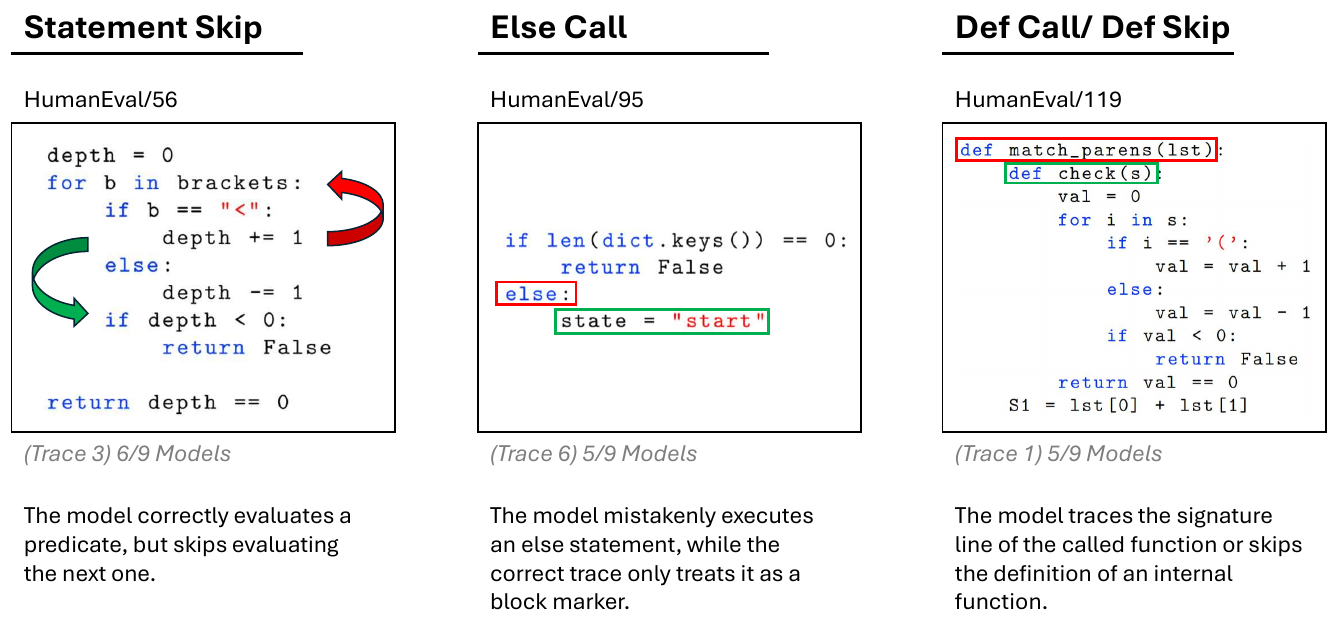} % Fit to width
\caption{Examples of three of the most common error types encountered in HumanEval tracing. The subtitle denotes the number of models that exhibited this error (as the first deviation) in a given Trace.}
\label{fig:Errors}
\end{figure*}

In order to better understand the types of errors produced by models when evaluating program calls, we conducted a qualitative study by manually reviewing a set of ten randomly selected HumanEval traces for each of the nine models. Among these, 23\% of the traces were identified correctly. 
We observed several common error patterns across all models, which we outline in \autoref{fig:Errors}.
Generally, models struggled with trace-specific issues, often arising from particular statements, such as `else` statements, predefined functions, and similar constructs that serve as block markers but are not executed by the program. These cases accounted for 20\% of the 90 analyzed traces. Notably, errors were distributed equally across different model sizes, with the exception of the definition call, which was explicitly described in the prompt. Simpler mistakes, such as skipping statements or conditions, accounted for 16\% and 8\% of the cases, respectively. Statement skips, as demonstrated in \autoref{fig:Errors}, were particularly frequent in the given HumanEval problem, during which models failed to execute the depth check before advancing to the next loop iteration. This issue occurred predominantly when the depth was greater than or equal to zero, possibly influencing the model to skip the check. Condition skips, where the body of a conditional statement was executed without evaluating the predicate, seem to result from similar model behavior. A related error type, which we classified as "loop skip," accounted for 4\% of the cases. This error occurs when a model exits a loop prematurely.

We also identified a group of error types more common in smaller, less proficient models. These included the ``nonsense'' error (12\%), which refers to accessing lines that are either empty or contain only comments, the ``empty'' error (9\%), where the model failed to produce a valid list, and the ``predicate wrong'' error (6\%), where an incorrectly evaluated predicate led to an incorrect control flow jump. Lastly, the ``no exit'' error occurred when the model produced an unclosed list. Some models, such as CodeLLama, struggled with format expectations, often producing empty or non-sensical outputs.

We further evaluated five random traces for each of the advanced topics using the best-performing model, Gemini 1.5-Pro. Surprisingly, the most frequent errors encountered were not specific to the corresponding topic. A majority of the traces involved definition call and skip errors, which were more prominent in this context compared to HumanEval, likely due to the multiple functions present in the programs. One trace, which focused on object-oriented principles, exhibited an error in resolving a nested expression, where the inner object initialization was simply skipped. Similarly, a concurrency trace displayed an error related to the marking of concurrency. We note that such errors likely also occur in the other traces, but they were not the first ones, which we limited our \linebreak analysis to.

In general, the advanced nature of concept-based programs makes it more challenging for models to accurately represent program execution. This is due to the increased length and complexity of the traces, as well as the nested structure of the programs, which include multiple functions or class definitions. When considering realistic software suites, which contain exponentially more lines of code and significantly more complex control flows between different files and packages, it is clear that current state-of-the-art models are still far from being able to fully capture the functionalities of realistic software systems.

\subsection{Alternative Prompts and Settings}
To validate our current elicitation approach, we experiment with alternative prompt techniques and line annotations.  We apply these techniques on a small subset of one test case for every task ID, resulting in 163 samples. The results are shown in Table~\ref{tab:prompting_techniques}. We show CoT as our baseline and first experiment with static analysis. A lot of the programs contain output or intermediate variables that do not have an influence on the control flow. We ask the model to seek out the variables that do have an influence and disregard the other ones. In addition, we ask the model to try and determine the number of loop iterations before entering and simply reproducing the steps the extracted number of times rather than actually executing it. We found that the model only applies limited analysis, sometimes even ignoring the instruction. Correspondingly, we observe a small performance decrease. We also experiment with increasing the number of samples given in the prompt. In particular, we add one code sample detailing a recursive program execution, and one more that contains nested method calls. We observe no significant performance increase, showing that the model generally seems to possess knowledge of the demonstrated execution concepts, only requiring one prompt to learn the correct output format. Apart from the prompt, we test alternatives to simply annotating the line numbers as comments. Since LLMs are known to struggle with numerical representations, we use a combination of singular letters and symbols, as well as the word \textbf{Line}, prior to the index. Both of these approaches reduced the performance, indicating that line indices are a sufficient annotation choice.

As denoted in earlier sections, we generally employ Greedy-decoding, as we assume it to be more applicable to a ground-truth task such as execution tracing. However, to validate these assumptions, we regenerate the samples of the best model, Gemini, using an increased temperature of 1.0. Throughout the CoT and Direct prompting, we observe a performance decrease of about 3 percent average accuracy for Chain-of-Thought and a 10 percent decrease for direct prompting.

\begin{table}[!th]
\centering
\caption{Comparison of different Prompting Techniques using Gemini1.5-Pro 002  on a HumanEval subset}
\label{tab:prompting_techniques}
\begin{tabular}{@{}lcc@{}}
\toprule
\textbf{Prompting Technique} & \textbf{Average Acc} & \textbf{Average Sim} \\ \midrule
CoT                          & \textbf{0.6522}                    & 0.8914                      \\ 
Static Analysis              & 0.6149                    & 0.8843                      \\ 
3-Shot                       & 0.6522                    & \textbf{0.8967}                      \\ 
Symbols                      & 0.0311                    & 0.7787                      \\ 
Line                         & 0.5776                    & 0.8780                      \\ \bottomrule
\end{tabular}

\end{table}

\section{Related Work}
\mypara{Coding Benchmarks for LLMs} Numerous coding benchmarks have been proposed for LLMs in recent years. We summarize and compare a few popular ones here. Open AI HumanEval and MBPP~\cite{austin2021programsynthesislargelanguage} are the most popular code generation benchmarks. Several other benchmarks such as CruxEval~\cite{gu2024cruxeval}, LiveCodeBench~\cite{jain2024livecodebench}, and Codemind~\cite{liu2024codemind} include execution reasoning tasks (in addition to code generation) such as predicting the output of a code snippet given some input. 
CodeXGlue~\cite{lu2021codexglue} is another prior dataset consisting of 10 different coding tasks. However, it does not include any execution-related tasks.

Runtime Reasoning REval~\cite{chen2024reasoning} is the most recent work that proposed four execution-related tasks: coverage prediction, program state prediction, execution path prediction (prediction next statement), and output prediction. However, these tasks do not require the LLMs to reason about control flow or different programming structures like the tasks in \tool.
% CruxEval~\cite{gu2024cruxeval} includes input and output prediction tasks for 800 small Python functions (3-13 lines). In addition to code generation, LiveCodeBench~\cite{jain2024livecodebench} includes execution reasoning tasks that require predicting the output of code or tests when given the code. Codemind~\cite{liu2024codemind}  includes an 

Ma \etal~\cite{ma2024lmsunderstandingcodesyntax} evaluate LLMs on many different tasks, including generating Abstract Syntax Trees, Control Flow, and Call Graphs. However, they limit their investigation of the dynamic behavior of the execution to Equivalent Mutant Detection and Flaky Test Reasoning, which does not directly concern structural understanding abilities. 

Hooda \etal~\cite{hooda2024largecodemodelsunderstand} show that LLMs are vulnerable to different mutations related to control and data flow, as well as type and identifier assignment. A strong understanding of the full execution trace would help build resilience against such approaches. To the best of our knowledge, there is no prior work on evaluating the execution tracing abilities of LLMs, which we investigate in this work.

\mypara{Training LLMs for better execution reasoning}
Few recent approaches focused on improving the execution reasoning of LLMs.
 For instance, Ding \etal developed a new coding dataset augmented with tests and execution traces and trained an LLM, called  SemCoder~\cite{ding2024semcoder}. They showed that such a training strategy elicits better code generation and execution reasoning from the LLM. We also tested SemCoder on CoCoNUT, however, the model was not able to correctly follow the natural language instructions, resulting in no meaningful results.
% Mercury-code efficiency
% ~\cite{du2024mercury}
% Codemind~\cite{liu2024codemind}
% R2E~\cite{jain2024r2e}
%Runtime Reasoning REval~\cite{chen2024reasoning} (seems very related)
%
% Even though the application of large language models to code is a rather young discipline, 
Ni \etal\cite{ni2024nextteachinglargelanguage} showed that fine-tuning LLMs on Chain-of-Thought reasoning over execution trace improved the performance of PaLM on the HumanEval and the MBPP benchmarks~\cite{austin2021programsynthesislargelanguage}. However, their approach to tracing mostly consists of variable states for straight-line code, which they insert into the source code as comments instead of control flow reasoning. While they demonstrate that this approach also works without inserting the trace, they note that the model exhibits hallucination issues while adapting the trace into natural language reasoning steps. This naturally motivates enhancing language models' abilities to directly extract execution representations. 
%Another motivating factor for this work

\section{Conclusion and Future Work}
In this paper, we have introduced CoCoNUT, a benchmark for evaluating a model's capability to trace the control flow of a program given relevant input. Our evaluation of state-of-the-art models reveals significant gaps in their performance compared to generative tasks. We observe a strong correlation between trace length and performance, with models generally struggling to trace executions longer than 10 lines—a capability that appears to emerge only at larger scales. Notably, specific training on code does not consistently outperform increases in parameter size. Models uniformly fail on advanced structural programs with longer base traces, highlighting their limitations in predicting the full execution of complex code despite their widespread use in software engineering.

We also investigated the effects of prompting techniques. Chain-of-Thought (CoT) prompting generally improves performance, particularly for smaller models, but offers only marginal benefits for larger ones and sometimes propagates errors, leading models to incorrectly pursue alternative traces. While we experimented with more sophisticated prompt engineering on a subset of HumanEval, we did not find significant improvements over CoT.

To explore the connection between Code Generation and Execution Tracing, we used HumanEval as a proxy. Despite strong performance in code generation, tracing HumanEval solutions yielded significantly lower results, with performance only slightly higher than half the corresponding pass rate on generation. Additionally, we found no significant correlation between task and trace difficulty for the two approaches. This indicates that models lack the human-like ability to trace code they generate, raising concerns about their role in software development.

\mypara{Next Steps and Future Work}
An interesting next step, which we leave to future work, would be to conduct fine-tuning on our dataset, which could improve performance on adjacent tasks such as error localization or code summarization. This approach may help models differentiate between the functionality of code components and their natural language token equivalents, thereby enhancing their ability to reason about code behavior more effectively.
We intentionally adapted code snippets from sources that provide equivalent code in multiple programming languages, as we aim to expand our benchmark across them. We believe that structural code understanding is the most significant if it is achieved across several different languages since this implies nuanced understanding rather than pure knowledge of when to use certain tools or commands in one specific language. We also publish our relevant repository to enable the testing of a wider array of models, as well as potential training or fine-tuning on the execution trace data. 

% \section{Next Steps}
% In the next steps of the project, we will investigate the effects of different prompting techniques, like the employment of Chain-of-Thought. One key idea that we have developed is to drive the model to focus on relevant state variables while abandoning those variables and calls which do not influence the tracing. Such a behavior could be reached by an initial static analysis for example. We visualize the distinction between relevant and irrelevant parts of the code in \ref{fig:2}. [REMOVE?] Similarly, we explore the effect of the number of shots given in the prompt since efficient in-context learning has been shown to be competitive with full-on fine-tuning (\cite{yin2024deeperinsightsupdatespower}). Besides, we will test the capabilities of smaller, as well as code-specific models like DeepSeek-Coder V2 \cite{zhu2024deepseek} and QwenCoder 2.5 \cite{qwen}.

\bibliography{references}
\bibliographystyle{IEEEtran}

\newpage

\appendix

\subsection{Trace Length Buckets}

In the following, we display the measured accuracies by trace-lengths per the outlined buckets. 

\begin{table}[!htb] 
    \centering
    \setlength{\tabcolsep}{0.5em}
    \caption{Model Accuracy by Buckets of Trace Length using Direct Prompting.}
    \label{tab:TraceBuckets}
    \begin{tabular}{@{}lccccccc@{}}
        \toprule
        \textbf{Model}   & \textbf{1-3} & \textbf{3-5} & \textbf{5-10} & \textbf{10-25} & \textbf{25-40} & \textbf{40+} \\ \midrule
        Gemini1.5-Pro 002   & 93.8\% & \textbf{91.0\%} & \textbf{84.8\%} & \textbf{56.2\%} & \textbf{24.0\%} & \textbf{2.0\%} \\
        Claude3.5-Sonnet  & \textbf{98.6\%} & 86.6\% & 65.0\% & 26.6\% & 22.9\% & 0.7\% \\
        GPT4o      & 76.6\% & 62.7\% & 44.6\% & 12.0\% & 3.1\% & 0.0\% \\
        LLama3.1 70B & 81.7\% & 59.7\% & 19.2\% & 1.3\% & 0.0\% & 0.0\% \\
        CodeLLama 34B & 19.0\% & 25.4\% & 12.4\% & 0.0\% & 0.0\% & 0.0\% \\
        Qwen2.5-Coder 32B & 96.2\% & 63.4\% & 24.9\% & 0.9\% & 0.0\% & 0.0\% \\
        Codestral 22B & 33.5\% & 42.5\% & 15.8\% & 0.4\% & 0.0\% & 0.0\% \\
        LLama3.1 8B & 20.0\% & 27.6\% & 7.3\% & 0.0\% & 0.0\% & 0.0\% \\
        Qwen2.5-Coder 7B & 7.9\% & 9.0\% & 3.4\% & 0.0\% & 0.0\% & 0.0\% \\
        CodeLLama 7B & 0.0\% & 0.0\% & 0.0\% & 0.0\% & 0.0\% & 0.0\% \\
        \bottomrule
    \end{tabular}
    
    \vspace{0.2cm}
\end{table}

\begin{table}[!htb]
\centering
\caption{Model Accuracy Across Trace Length Buckets using Chain-of-Thought prompting.}
\setlength{\tabcolsep}{0.5em}
\label{tracebuckets2}
\begin{tabular}{@{}lccccccc@{}}
\toprule
\textbf{Model} & \textbf{1-3} & \textbf{3-5} & \textbf{5-10} & \textbf{10-25} & \textbf{25-40} & \textbf{40+} \\ \midrule
Gemini1.5-Pro 002  & 95.5\% & \textbf{88.8\%} & \textbf{84.8\%} & \textbf{54.9\%} & \textbf{26.0\%} & \textbf{1.3\%} \\
Claude3.5-Sonnet  & \textbf{97.2\%} & 85.8\% & 71.8\% & 41.2\% & 30.2\% & 0.7\% \\
GPT4o & 63.5\% & 83.6\% & 60.5\% & 16.7\% & 9.4\% & 0.0\% \\
LLama3.1 70B & 72.4\% & 67.9\% & 46.9\% & 8.2\% & 4.2\% & 0.0\% \\
CodeLLama 34B & 15.2\% & 17.2\% & 8.5\% & 0.0\% & 0.0\% & 0.0\% \\
Qwen2.5-Coder 32B & 84.5\% & 74.6\% & 50.3\% & 9.4\% & 0.0\% & 0.0\% \\
Codestral 22B & 48.3\% & 56.0\% & 24.9\% & 1.3\% & 1.0\% & 0.0\% \\
LLama3.1 8B & 21.0\% & 38.8\% & 19.8\% & 0.0\% & 0.0\% & 0.0\% \\
Qwen2.5-Coder 7B & 24.8\% & 30.6\% & 2.3\% & 0.0\% & 0.0\% & 0.0\% \\
CodeLLama 7B & 0.0\% & 0.1\% & 0.0\% & 0.0\% & 0.0\% & 0.0\% \\
\bottomrule
\end{tabular}
\end{table}

\subsection{HumanEval Solution Line Length}

\begin{figure}[!htb]
    \centering
    \includegraphics[width=\columnwidth]{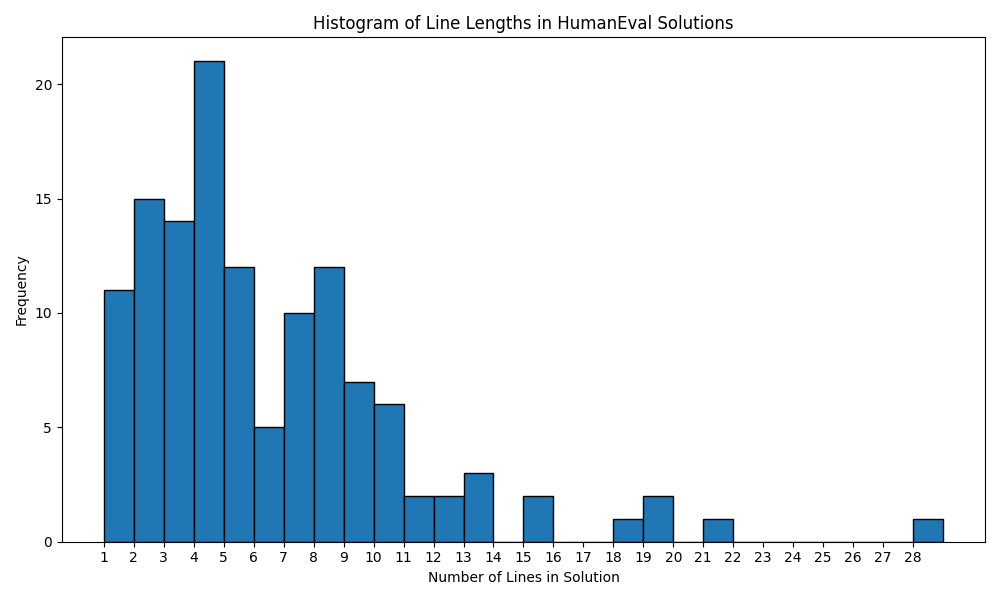}
    \caption{Length distribution across HumanEval solutions.}
    \label{fig:HumanEval_length}
\end{figure}

\subsection{Prompts}

\newsavebox{\firstlisting}
\begin{lrbox}{\firstlisting}
\begin{minipage}{0.45\textwidth}  % Adjust the width of the box
\begin{lstlisting}[language=Python, numbers=none, frame=none]
def simple_loop(x):  #1
    for i in range(3):  #2
        print(i + x)  #3
    return i  #4
\end{lstlisting}
\end{minipage}
\end{lrbox}

We include a basic prompt for the CoT approach on the HumanEval Trace task in the following (note that the contained code usually contains appropriate newlines and indentation):

\lessonblock{
This task will evaluate your ability to appreciate the control flow of code with a given input.
In the following, I will give you the source code of a program written in Python. The program may feature different functions, which may call each other.
To make the task more accessible to you, I have annotated the lines with their index as comments (those begin with a \#).
The following is very important! Please note that the function signatures are generally not called,
instead you should start with the first line of the function. This does not apply to the function call, of course.
In addition to the function, I will give you an initial input and the called function.
It is your task to return the called lines, in order, as a list. I will give you an example:\newline
\usebox{\firstlisting}\newline
Input: (5) \newline
Correct solution: [2,3,2,3,2,3,2,4] \newline
Now I will give you your task.
Here is the source code: [code]\newline
Here is the called function: [function]\newline
Here is the input to the function [input]\newline
Please produce the python list containing the executed line numbers in order now. Remember not to include the function signature lines. Think about the solution step-by-step,
going through execution steps one at a time. Finally, print the solution as a list of executed steps.}

The prompts for the advanced topics are adjusted to account for the difference in format. The prompts for concurrency also feature an adapted code example.

\end{document}